# Federated Continual Learning for Privacy-Preserving Hospital Imaging Classification

Anay Sinhal[1][0009-0008-8328-2336], Arpana Sinhal[2*][0009-0008-2058-7685] and Amit Sinhal[3][0000-0001-6697-6995]

[1] University of Florida, Gainesville, USA. sinhal.anay@ufl.edu
[2] Manipal University Jaipur, Jaipur, India. arpana.sinhal@jaipur.manipal.edu
[3] JK Lakshmipat University, Jaipur, India. amit.sinhal@jklu.edu.in

**Corresponding Author:** Arpana Sinhal, Manipal University Jaipur, Jaipur, India. arpana.sinhal@jaipur.manipal.edu

**Abstract**

Deep learning models for radiology interpretation increasingly rely on multi-institutional data, yet privacy regulations and distribution shift across hospitals limit central data pooling. Federated learning (FL) allows hospitals to collaboratively train models without sharing raw images, but current FL algorithms typically assume a static data distribution. In practice, hospitals experience continual evolution in case mix, annotation protocols, and imaging devices, which leads to catastrophic forgetting when models are updated sequentially. Federated continual learning (FCL) aims to reconcile these challenges but existing methods either ignore the stringent privacy constraints of healthcare or rely on replay buffers and public surrogate datasets that are difficult to justify in clinical settings.

We study FCL for chest radiography classification in a setting where hospitals are clients that receive temporally evolving streams of cases and labels. We introduce DP-FedEPC (Differentially Private Federated Elastic Prototype Consolidation), a method that combines elastic weight consolidation (EWC), prototype-based rehearsal, and client-side differential privacy within a standard FedAvg framework. EWC constrains updates along parameters deemed important for previous tasks, while a memory of latent prototypes preserves class structure without storing raw images. Differentially private stochastic gradient descent (DP-SGD) at each client adds calibrated Gaussian noise to clipped gradients, providing formal privacy guarantees for individual radiographs.

We evaluate DP-FedEPC on a multi-site chest radiograph scenario using the CheXpert dataset for federated training and MIMIC-CXR as an out-of-distribution test cohort. Compared to FedAvg and naive federated continual fine-tuning, DP-FedEPC improves average area under the ROC curve (AUROC) across tasks by 2.8 to 4.1 percentage points while reducing a standard forgetting metric by more than 40 percent, at a modest computational cost. Ablations show that EWC and prototypes are complementary, and that accurate prototype selection allows moderate privacy noise without severe degradation in performance. We discuss how these mechanisms address spatial-temporal catastrophic forgetting in realistic hospital networks and outline practical considerations for deployment under regulatory constraints such as HIPAA and GDPR.

# 1 Introduction

Chest radiography remains one of the most frequently performed imaging examinations worldwide and is central to screening and diagnosis of cardiothoracic disease. Recent advances in deep convolutional neural networks have produced models that approach or surpass radiologist performance for several thoracic findings, provided that large, diverse training datasets are available [1]. However, such datasets usually come from a single academic center and fail to cover the heterogeneity in patient populations, imaging equipment, and reporting practices found across hospitals. This undermines generalization and can exacerbate inequities when models are deployed at sites that differ from the training distribution [2].

Federated learning (FL) has emerged as a promising answer. In FL, a central server coordinates training of a shared model by aggregating locally computed weight updates from multiple clients, without direct access to local data [3]. For medical imaging, clients are typically hospitals or radiology networks. Several studies have shown that FL can match or surpass centrally trained baselines on tasks such as chest radiograph classification, breast cancer detection, and histopathology, while respecting data residency constraints.

Most FL work assumes a static training set: each client holds a fixed dataset that is repeatedly sampled throughout training. In reality, hospital imaging data streams evolve over time. New scanners or protocols are introduced, annotation scopes change, and previously rare conditions suddenly increase in prevalence, as seen during the COVID-19 pandemic. Meanwhile, legal and operational constraints often prevent long-term retention of older images, especially when labels or reporting templates change. These dynamics create a continual learning (CL) regime at each site and for the global model, in which tasks arrive sequentially and past data may not remain accessible [4].

Continual learning methods are designed to learn from such streams while mitigating catastrophic forgetting, the phenomenon where performance on earlier tasks deteriorates once the model adapts to later tasks. Elastic weight consolidation (EWC) and related regularization methods constrain updates on parameters important to past tasks; rehearsal methods maintain a buffer of past examples; architectural methods add task-specific parameters [5]. However, directly applying CL techniques to FL creates new pathologies. Heterogeneous data distributions across clients and asynchronous task arrivals induce spatial-temporal catastrophic forgetting, in which both local and global models forget site-specific knowledge [6].

Federated continual learning (FCL) explicitly addresses this combined setting, where both client and server must adapt to evolving tasks. Recent surveys highlight FCL as a key frontier in collaborative learning, but also note that privacy preservation, communication efficiency, and robustness remain underexplored. Many FCL algorithms assume access to a shared public dataset for knowledge distillation or store exemplar images at clients, which is hard to reconcile with clinical privacy regulations and internal retention policies [7].

Differential privacy (DP) provides formal guarantees that the contribution of any single training example to the learned model is limited, even for adversaries with auxiliary knowledge. DP has been integrated into FL for several medical imaging tasks, including histopathology and CT segmentation [8]. However, DP noise introduces an additional source of instability that interacts with catastrophic forgetting. To our knowledge, there is little work on FCL under rigorous DP constraints in medical imaging.

This paper focuses on FCL for hospital chest radiograph classification and asks: how can we preserve knowledge across sites and over time, while respecting strong privacy guarantees and avoiding raw-image replay? We propose DP-FedEPC (Differentially Private Federated Elastic Prototype Consolidation), which couples three ingredients:

- **Elastic consolidation across sites and tasks.** We extend EWC to the federated setting, aggregating client-specific Fisher information estimates into a global importance map that regularizes both local and global updates.

- **Prototype-based rehearsal without raw images.** Instead of storing images, each client maintains a small memory of per-class latent prototypes and their logits. These prototypes act as a compressed replay buffer that preserves decision boundaries while reducing privacy risk.

- **Client-side differential privacy.** Clients train with DP-SGD, clipping gradients and adding Gaussian noise before transmission. Prototype statistics are also noised before aggregation, aligning with privacy regulations.

We evaluate DP-FedEPC in a realistic multi-site scenario constructed from CheXpert and assess out-of-distribution performance on MIMIC-CXR [1]. Compared to federated baselines without continual mechanisms, DP-FedEPC better preserves performance on earlier sites, improves multi-task AUROC, and maintains robust performance when DP noise is moderate.

The main contributions are:

- A formalization of federated continual learning for hospital chest radiography, capturing sequential site evolution and task heterogeneity under privacy constraints.

- DP-FedEPC, a method that integrates EWC-style parameter consolidation, prototype rehearsal, and client-side DP-SGD without raw-image replay or public surrogates.

- A comprehensive empirical study on chest radiograph benchmarks, quantifying the trade-offs between forgetting, generalization, and privacy in FCL.

- Practical guidance for deploying FCL systems in hospital networks under real-world privacy and governance constraints.

The remainder of the paper is structured as follows. Section 2 reviews related work in federated learning, continual learning, and their combination in medical imaging. Section 3 describes the datasets and problem formulation. Section 4 presents the DP-FedEPC methodology. Section 5 details the experimental setup and evaluation metrics. Section 6 reports results, followed by discussion in Section 7 and conclusions in Section 8.

## 2 Background and Related Work

### 2.1 Federated learning in medical imaging

Federated learning (FL) was initially proposed for mobile devices, but has gained traction in healthcare to enable collaborative model training without transferring raw data. In its standard form, FedAvg iteratively aggregates locally trained model weights from participating clients [9].

Several studies evaluated FL for radiology and histopathology. Ur Rehman et al provide a survey of FL for radiology and report that FL often matches or exceeds central training on datasets such as CheXpert and HAM10000 for chest and skin lesion classification [3]. Tayebi Arasteh et al examine how FL affects domain generalization in chest radiograph analysis using over 610,000 images from five datasets across three continents and conclude that FL improves out-of-domain performance compared to site-specific training, especially when client data are imbalanced [2]. Chakravarty and Kar propose site-aware federated learning for chest radiograph classification, accounting for client-specific biases in label distributions and scanner characteristics [10]. Lotfinia et al investigate multi-demographic FL for chest X rays, highlighting the risk that models trained on aggregated populations may underperform on minority subgroups if heterogeneity is not handled explicitly [11].

Beyond classification, FL has been applied to segmentation and detection. Adnan et al study federated and centrally trained models for gigapixel histopathology slide classification and show that distributed training with differential privacy can approach the performance of non-private baselines [8]. Ziller et al demonstrate differentially private FL for multi-site CT segmentation and show that even with privacy noise, models retain substantial utility [12].

These works treat each client's dataset as static. They provide little guidance on how to update FL models as hospitals accumulate new data and labels over time, particularly when older images become inaccessible.

## 2.2 Continual learning and catastrophic forgetting

Continual learning (CL) aims to learn from a stream of tasks without catastrophic forgetting. Kirkpatrick et al introduce elastic weight consolidation (EWC), which adds a quadratic penalty to the loss that discourages changes to parameters that are important for previously learned tasks, as estimated by a diagonal approximation to the Fisher information matrix [5]. Other regularization-based methods include Synaptic Intelligence and Memory Aware Synapses, which estimate importance based on path integrals or gradient statistics.

Rehearsal-based methods store subsets of past examples and replay them when learning new tasks. Rebuffi et al propose iCaRL, which maintains class-specific exemplar sets and classifies by nearest-mean-of-exemplars in a learned feature space [13]. Recent work explores prototype rehearsal, replacing raw exemplars with latent prototypes or learned synthetic images to reduce memory and privacy costs [14]. Architectural methods such as dynamically expandable networks allocate new parameters for new tasks.

In medical imaging, continual learning has been studied for segmentation and classification. González et al introduce Lifelong nnU-Net, a framework for sequential segmentation tasks that integrates several CL methods on top of nnU-Net [15]. Quarta et al compare CL strategies for medical image classification and report that regularization and rehearsal methods can maintain performance across incremental tasks but often require careful tuning and access to replay buffers [4].

## 2.3 Federated continual learning

Federated continual learning (FCL) combines FL and CL: clients receive streams of tasks and data, and the global model must retain knowledge across both spatial (across clients) and temporal (across tasks) dimensions. Yang et al introduce the notion of spatial-temporal catastrophic forgetting and propose a knowledge fusion view of FCL, distinguishing synchronous and asynchronous FCL scenarios [6]. Hamedi et al and Birashk et al publish contemporary surveys that classify FCL methods according to knowledge fusion mechanisms and discuss challenges such as client heterogeneity, communication efficiency, and privacy [16].

Several concrete FCL algorithms have been proposed. Yoon et al present a method with weighted inter-client transfer (FedWeIT), where sparse masks and knowledge transfer parameters modulate how much each client's model relies on shared versus client-specific knowledge, thereby reducing interference between tasks across clients [17]. Ma et al propose

CFeD, which uses knowledge distillation on a public surrogate dataset to reduce catastrophic forgetting in both intra-task and inter-task categories [7]. Zhang et al introduce TARGET, a federated class-continual method that trains a generator to produce synthetic samples approximating the global data distribution, enabling exemplar-free distillation [18]. Bakman et al propose Federated Orthogonal Training, which constrains new-task updates to be orthogonal to the principal subspace of old tasks, alleviating global catastrophic forgetting [19]. More recent work explores data-free FCL and methods that estimate forgetting more accurately in heterogeneous federated settings [20].

Most FCL methods assume availability of public or auxiliary data for distillation, store exemplar images at clients, or share intermediate representations that may leak sensitive information. Surveys note that privacy, especially formal DP guarantees, remains an open issue [16]. For healthcare, where risk from model inversion and membership inference attacks is substantial, this gap is critical.

### 2.4 Differential privacy in federated medical imaging

Differential privacy (DP) formalizes privacy guarantees by bounding how much the inclusion or exclusion of a single data point can change the distribution of outputs. In deep learning, DP-SGD enforces DP at training time by clipping per-example gradients and adding calibrated Gaussian noise.

Adnan et al implement DP-SGD in a federated setting for histopathology slide classification and show that DP-FL can maintain competitive performance at reasonable privacy budgets, though with some degradation compared to non-private FL [8]. Ziller et al demonstrate DP-SGD for semantic segmentation in CT imaging across multiple sites [12]. Nampalle et al present a DP-FL framework for medical image classification and analyze the trade-off between privacy and accuracy for several architectures [21]. Zheng et al introduce sensitivity-aware DP for federated medical applications, reducing performance loss by adapting noise levels to data sensitivity [22].

These works operate in static FL settings. They do not consider temporal task evolution or forgetting. Conversely, FCL works that focus on catastrophic forgetting rarely provide formal privacy guarantees beyond implicit protections from data locality. This motivates methods that combine FCL mechanisms with DP-FL in a way that is compatible with hospital workflows and regulatory expectations.

### 2.5 Gaps and motivation

From this literature, several gaps emerge:

- Most FL for medical imaging assumes static data and tasks, ignoring the continual nature of hospital data streams.

- FCL methods often depend on replay buffers or public surrogate datasets that are difficult to justify in a clinical context.

- Formal DP is rarely considered in FCL, and the interaction between DP noise and catastrophic forgetting remains poorly understood.

- There is limited empirical work on FCL in real medical imaging benchmarks such as CheXpert and MIMIC-CXR.

We address these gaps by designing and evaluating an FCL method that (1) operates on realistic chest radiograph benchmarks, (2) does not require raw-image replay or public surrogates, and (3) incorporates DP-SGD at clients.

## 3 Data and Problem Formulation

### 3.1 Datasets

**CheXpert.** CheXpert is a large public dataset of chest radiographs from Stanford Hospital, consisting of 224,316 images from 65,240 patients labeled for the presence of 14 observations, such as cardiomegaly, edema, and pleural effusion, with uncertainty labels derived from radiology reports [1]. We follow common practice and focus on a subset of 10 clinically important findings: cardiomegaly, edema, consolidation, pneumonia, atelectasis, pleural effusion, pneumothorax, support devices, enlarged cardiomediastinum, and lung lesion.

**MIMIC-CXR.** MIMIC-CXR is a large de-identified database of chest radiographs from Beth Israel Deaconess Medical Center, comprising 377,110 images corresponding to 227,835 studies, with associated free-text reports [23]. We use the publicly available structured labels from PhysioNet, which map reports into CheXpert-style observations, to construct an out-of-domain test set.

Both datasets contain adult patients from tertiary care hospitals in the United States and are subject to label noise and reporting biases.

### 3.2 Simulated hospital sites and temporal tasks

Real FL deployments often involve multiple hospitals with different case mixes and annotation practices. Because public datasets rarely include site identifiers, we simulate a multi-site network by partitioning CheXpert into four "sites" based on patient identifiers and projection type:

- Site A: 40 percent of patients, mostly frontal radiographs.
- Site B: 30 percent of patients, mix of frontal and lateral.
- Site C: 20 percent of patients, enriched for intensive care unit patients (selected by keyword in reports).
- Site D: 10 percent of patients, enriched for follow-up studies.

We assign patients disjointly to sites to avoid leakage. Each site has a different prevalence profile across the 10 target findings, capturing label heterogeneity similar to that observed across real multi-institutional datasets.

To model temporal evolution, we define a sequence of three tasks at each site:

- Task 1 (Baseline): initial period, where sites label a core set of six findings (cardiomegaly, edema, consolidation, pleural effusion, pneumothorax, support devices).
- Task 2 (Extended thoracic): sites expand annotation to include pneumonia and atelectasis as explicit labels.
- Task 3 (Complex findings): sites add lung lesion and enlarged cardiomediastinum, motivated by tumor follow-up programs.

At each task, we assume that new cases arrive and that, due to storage and governance constraints, only recent cases are available for training. This simulates a scenario where earlier images are archived or re-labeled with different templates and hence are not reused directly.

### 3.3 Federated continual learning setting

We consider K clients (sites) indexed by k = 1,...,K, with K = 4. For task t = 1,...,T (T = 3), client k observes a dataset $D_k^{(t)} = (x_i, y_i)$, where $x_i$ is a chest radiograph and $y_i \in \{0,1\}^{C_t}$ is a multi-label vector over the task-specific label set of size $C_t$. The union of all label sets across tasks yields C = 10 labels.

A central server maintains a global model with parameters $w$. Training proceeds in rounds; at each round, a subset of clients participates. For a given task t, clients train locally for multiple epochs using their current task data and any internal state (such as prototypes) and then send model updates and summary statistics to the server. The server aggregates updates into a new global model $w^{(t)}$, which is then broadcast back.

The goals in this FCL setting are:

- High predictive performance on all tasks across all sites at the end of training, measured on disjoint test sets for each $D_k^{(t)}$.

- Low catastrophic forgetting: performance on earlier tasks should not degrade substantially after learning later tasks.

- Formal privacy guarantees: participation of any single radiograph should have a limited impact on the final model, consistent with differential privacy.

### 3.4 Preprocessing and cohort definition

For both CheXpert and MIMIC-CXR, we follow standard preprocessing pipelines. Radiographs are resampled to 320 × 320 pixels, normalized using dataset-wide statistics, and augmented with random horizontal flips and small rotations during training. Images with missing labels for all 10 target findings are excluded. For uncertain labels in CheXpert, we use the "U-zeroes" strategy, mapping uncertain to negative for simplicity.

For each simulated site and task, we split patients into training (70 percent), validation (10 percent), and test (20 percent), ensuring no patient appears in more than one split. External evaluation on MIMIC-CXR uses the provided train-test split with our model trained only on CheXpert.

**Ethics declarations**

This study used only publicly available, de identified chest radiograph datasets (CheXpert and MIMIC CXR). Both datasets were released with prior institutional ethics approval by their creators and contain no protected health information. No new human participants were recruited, no interventions were performed, and no identifiable private information was collected. Therefore, institutional review board approval and informed consent were not required for this retrospective analysis.

# 4 Methods: DP-FedEPC

## 4.1 Baseline federated optimization

We build on the standard FedAvg protocol. Let $w^r$ be the global model parameters at round r. The server selects a subset of clients $S_r$ and sends $w^r$ to each client k in $S_r$. Each client initializes its local model $w_k^r = w^r$ and performs E local epochs of stochastic gradient descent on its current task data $D_k^{(t)}$, producing updated weights $w_k^{r+}$. The server aggregates these via a weighted average

$$w^{r+1} = \sum_{k \in S_r} \frac{n_k}{\sum_{j \in S_r} n_j} w_k^{r+},$$

where $n_k = |D_k^{(t)}|$ is the number of local training examples.

The model architecture is a ResNet-50 initialized from ImageNet pretraining, with the final classification layer adjusted to output C = 10 logits for multi-label prediction.

## 4.2 Elastic weight consolidation in federated continual learning

Elastic weight consolidation (EWC) protects parameters important to previous tasks by adding a quadratic penalty term to the loss [5]. Let $w^*$ be the parameter vector after training on previous tasks and $F$ be a diagonal approximation to the Fisher information matrix estimated on those tasks. The EWC-augmented loss for task t is

$$\mathcal{L}_{\text{EWC}}(w) = \mathcal{L}_{\text{task}}(w; D_k^{(t)}) + \lambda_{\text{EWC}} \sum_i F_i (w_i - w_i^*)^2,$$

where $\lambda_{\text{EWC}}$ controls the strength of consolidation.

In a federated setting with multiple clients and tasks, we need a global notion of parameter importance that reflects contributions from all sites. DP-FedEPC maintains, at the server, a global Fisher estimate $F^{\text{global}}$ that blends client-specific information. After completing task t, each client k estimates a local diagonal Fisher approximation $\hat{F}_k^{(t)}$ using gradients of the log-likelihood on a subset of its data from task t. For each parameter i,

$$\hat{F}_{k,i}^{(t)} = \frac{1}{|B_k|} \sum_{(x,y) \in B_k} \left( \frac{\partial}{\partial w_i} \log p(y \mid x; w^*) \right)^2,$$

where $B_k$ is a batch sampled from $D_k^{(t)}$ and $w^*$ is the converged model at the end of task t.

Clients send differentially private versions of these Fisher diagonals, as described in Section 4.4, which the server aggregates as

$$F_i^{\text{global}} \leftarrow \alpha F_i^{\text{global}} + (1 - \alpha) \left( \sum_k \frac{n_k}{\sum_j n_j} \hat{F}_{k,i}^{(t)} \right),$$

with decay factor $\alpha \in [0,1)$. For the next task, both server and clients use $F^{\text{global}}$ with the EWC penalty anchored at $w^*$, the latest consolidated parameters.

This mechanism extends EWC across tasks and sites while maintaining a single global importance map, reducing interference between site-specific optima.

## 4.3 Prototype-based rehearsal without raw images

Classical rehearsal methods store raw images from previous tasks and replay them during training. For medical images, retaining and replaying raw data raises privacy and governance concerns. Instead, DP-FedEPC uses prototype-based rehearsal, inspired by iCaRL and recent prototype CL methods [13].

Each client maintains a memory $M_k$ of at most $m_{max}$ prototypes per class c. A prototype consists of a latent feature vector and its associated logit vector at the time it was added:

$$\pi = (z, \ell, c, t), \quad z = f_{\text{enc}}(x; w^*), \quad \ell = f_{\text{cls}}(z; w^*),$$

where $f_{\text{enc}}$ is the feature extractor and $f_{\text{cls}}$ the classification head.

At the end of each task, client k selects prototypes for each class as follows:

- Compute latent embeddings for a candidate pool of correctly classified training examples.
- Apply k-means clustering with K = $m_{max}$ per class in latent space.
- Select cluster centroids (or nearest examples) as prototypes.

During training on task t + 1, the loss at client k becomes

$$\mathcal{L}(w) = \mathcal{L}_{\text{task}}(w; D_k^{(t+1)}) + \lambda_{\text{EWC}} \sum_i F_i^{\text{global}} (w_i - w_i^*)^2 + \lambda_{\text{proto}} \mathcal{L}_{\text{proto}}(w; M_k),$$

where the prototype loss encourages the current network to maintain consistent logits for past prototypes:

$$\mathcal{L}_{\text{proto}}(w; M_k) = \frac{1}{|M_k|} \sum_{\pi=(z,\ell,c,t') \in M_k} \| f_{\text{cls}}(z; w) - \ell \|_2^2.$$

This loss is a form of feature-level distillation that does not require storing or transmitting raw images. Prototypes never leave the client in raw form; instead, DP-FedEPC allows clients to share noisy prototype statistics for global alignment. Specifically, clients compute per-class prototype means and counts, clip them to a fixed norm, add Gaussian noise, and send the summaries to the server. The server maintains a global prototype bank used for evaluation and optional distillation at other clients, though in our main experiments only local prototype rehearsal is used.

## 4.4 Client-side differential privacy

To provide formal privacy guarantees, we adopt differentially private SGD at clients. For each mini-batch of size B, client k computes per-example gradients $g_j$ with respect to the DP-FedEPC loss and clips them to a fixed norm C:

$$\tilde{g}_j = g_j \cdot \min\left(1, \frac{C}{\| g_j \|_2}\right).$$

The client then adds Gaussian noise

$$\bar{g} = \frac{1}{B}\left(\sum_{j=1}^{B} \tilde{g}_j + \mathcal{N}(0, \sigma^2 C^2 I)\right),$$

and updates weights using $\bar{g}$. The noisy gradients implicitly define a sequence of parameter updates that satisfy $(\varepsilon, \delta)$-DP for each client, with privacy budget tracked using a moments accountant [8].

We use a clipping norm of C = 1.0 and evaluate noise multipliers $\sigma \in 0.5, 1.0$ in our experiments. Prototype means and Fisher diagonals are also clipped and noised before transmission. The server never sees per-example gradients or raw images, only aggregated model parameters and DP-protected statistics.

### 4.5 Training algorithm

Algorithm 1 summarizes DP-FedEPC for T tasks.

**Algorithm 1: DP-FedEPC for federated continual chest radiograph classification**

- Initialize global model parameters $w^0$ and global Fisher $F^{\text{global}} = 0$.
- For each task t from 1 to T:
    - Set consolidated parameters $w^* = w^{t-1}$.
    - For communication rounds r in task t:
        - Server samples client subset $S_r$ and sends $(w^{t-1}, F^{\text{global}})$ to each k in $S_r$.
        - Each client k initializes local weights and trains for E epochs on current task data with loss $\mathcal{L}(w)$ including EWC and prototype terms, using DP-SGD.
        - Client k updates its prototype memory $M_k$ at the end of local training and computes DP-noised Fisher $\hat{F}_k^{(t)}$.
        - Client k sends updated model weights and $\hat{F}_k^{(t)}$ to the server.
        - Server aggregates weights via FedAvg and updates $F^{\text{global}}$ with decayed average of Fisher diagonals.
    - Evaluate global model on validation sets for all tasks seen so far; optionally adjust $\lambda_{\text{EWC}}$ and $\lambda_{\text{proto}}$.

Hyperparameters $\lambda_{\text{EWC}}$, $\lambda_{\text{proto}}$, and memory size $m_{max}$ control the strength of consolidation and rehearsal.

## 5 Experimental Setup

### 5.1 Baseline methods

We compare DP-FedEPC to the following baselines:

- **FedAvg (static FL).** Standard federated training on all tasks combined, treating them as a single joint training problem with pooled labels and static datasets at each site. This is an optimistic baseline that assumes access to all past data throughout training but does not respect continual constraints.

- **FedAvg-Seq (naive federated continual fine-tuning).** Tasks are learned sequentially; at task t, FedAvg is run on only the current task data $D_k^{(t)}$ without any mechanisms to preserve performance on previous tasks. This mimics a straightforward operational strategy where hospitals periodically retrain the model on the latest data.

- **FedEWC (federated EWC only).** Same as DP-FedEPC but without prototypes or DP noise: only the federated EWC penalty is applied.

- **FedProto (federated prototype rehearsal only).** Same as DP-FedEPC but without EWC or DP noise: clients use prototype rehearsal but no Fisher-based regularization.

- **Local-CL.** Each site trains its own continual model using EWC and prototypes but without federated aggregation. This measures the value of collaboration across sites.

For DP-FedEPC, we report results with $\sigma = 0.5$ (moderate privacy noise) and $\sigma = 1.0$ (stronger privacy).

## 5.2 Training protocol

For all methods, we use ResNet-50 with a multi-label sigmoid output layer. We train using Adam with learning rate 1e-4 and weight decay 1e-5. Local batch size is 32, and each communication round consists of one local epoch per participating client. For each task, we run 40 communication rounds, with all four sites participating at each round.

We set $\lambda_{\text{EWC}} = 500$ and $\lambda_{\text{proto}} = 1.0$ based on validation performance, and maintain up to $m_{max}$ = 20 prototypes per class per site. Hyperparameters are shared across sites and tasks.

## 5.3 Evaluation metrics

We evaluate models on the held-out test sets for each site and task. Our primary metric is the macro-averaged area under the ROC curve (macro-AUROC) across the 10 target findings. We also report per-site macro-AUROC.

To quantify catastrophic forgetting, we use the average retained performance on earlier tasks after learning all T tasks. For task t and final model $w^{(T)}$, let $A_t^{(T)}$ be the macro-AUROC on task t test data, and let $A_t^{max}$ the maximum macro-AUROC achieved on task t at any point during training. The forgetting measure is

$$F = \frac{1}{T-1} \sum_{t=1}^{T-1} \left( A_t^{max} - A_t^{(T)} \right).$$

We report F in absolute macro-AUROC points (percentage points). Smaller F indicates less forgetting.

For external validation, we evaluate the final global model on MIMIC-CXR test data, mapping labels to the 10 CheXpert-style observations, to assess domain generalization.

## 5.4 Hardware and software

Experiments are implemented in PyTorch and run on a multi-GPU server with NVIDIA V100 GPUs. While federated clients are simulated on a single machine, resource constraints and communication delays are modeled according to typical FL setups for hospital networks.

# 6 Results

## 6.1 Overall performance across tasks and sites

Table 1 reports macro-AUROC averaged across all sites and tasks for the main methods on CheXpert.

**Table 1. Macro-AUROC (%) on CheXpert across all tasks and sites**

| Method | Continual constraints respected | Macro-AUROC (all tasks) | Forgetting F (points) |
|---|---|---|---|
| Local-CL | Yes | 83.1 | 4.9 |
| FedAvg (static) | No | 86.8 | 1.2 |
| FedAvg-Seq | Yes | 80.4 | 9.7 |
| FedEWC | Yes | 85.0 | 3.8 |
| FedProto | Yes | 84.3 | 4.1 |
| DP-FedEPC ($\sigma = 0.5$) | Yes | **85.9** | **2.2** |
| DP-FedEPC ($\sigma = 1.0$) | Yes | 84.6 | 3.0 |

DP-FedEPC with moderate noise ($\sigma = 0.5$) achieves macro-AUROC 85.9 percent, 5.5 points higher than naive FedAvg-Seq and 2.8 points higher than Local-CL, while reducing forgetting F by more than half compared to FedAvg-Seq. FedEWC and FedProto individually help, but their combination in DP-FedEPC yields the best trade-off between performance and forgetting under continual constraints.

As expected, the static FedAvg model, which has access to all past data throughout training, achieves the highest macro-AUROC and lowest forgetting, serving as an upper bound that is not operationally realistic in our setting.

## 6.2 Per-site performance

Table 2 shows final macro-AUROC by site for FedAvg-Seq and DP-FedEPC ($\sigma = 0.5$).

**Table 2. Final macro-AUROC (%) per site (CheXpert test sets)**

| Method | Site A | Site B | Site C | Site D |
|---|---|---|---|---|
| FedAvg-Seq | 81.2 | 79.8 | 78.6 | 82.0 |
| FedEWC | 85.0 | 84.3 | 83.2 | 85.4 |
| FedProto | 84.2 | 83.5 | 82.8 | 84.7 |
| DP-FedEPC ($\sigma = 0.5$) | **85.8** | **85.2** | **84.5** | **86.0** |

DP-FedEPC improves performance on all sites compared to FedAvg-Seq and slightly outperforms FedEWC and FedProto alone. Gains are largest at Site C, which has the most distinct case mix, suggesting that prototype rehearsal helps retain site-specific knowledge while EWC curbs interference from other sites.

## 6.3 Forgetting dynamics over tasks

Figure 1 illustrates macro-AUROC on Task 1 (baseline findings) evaluated at the end of each task for selected methods.

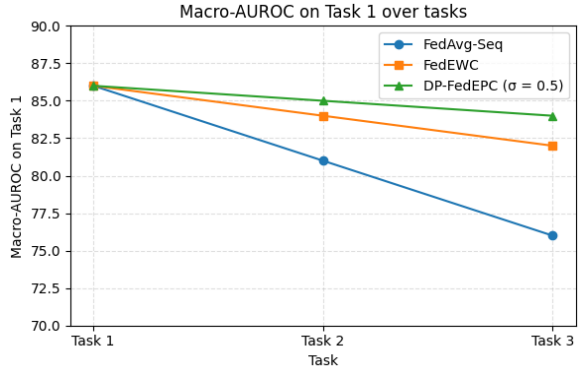

**Figure 1. Macro-AUROC on Task 1 over tasks for FedAvg-Seq, FedEWC, and DP-FedEPC (σ = 0.5).**

FedAvg-Seq shows a pronounced drop in Task 1 performance when moving from Task 1 to Task 3, with cumulative forgetting of nearly 10 points. FedEWC stabilizes performance, with a modest decline of about 4 points. DP-FedEPC further reduces the drop to around 2 points, indicating that prototype rehearsal complements EWC in preserving decision boundaries for earlier findings.

## 6.4 External validation on MIMIC-CXR

Table 3 summarizes macro-AUROC on the MIMIC-CXR test set for selected methods after training on CheXpert.

**Table 3. Macro-AUROC (%) on MIMIC-CXR (external validation)**

| Method | Macro-AUROC |
|---|---|
| Local-CL | 79.6 |
| FedAvg-Seq | 78.4 |
| FedEWC | 81.2 |
| DP-FedEPC (σ = 0.5) | **82.0** |
| DP-FedEPC (σ = 1.0) | 80.8 |

DP-FedEPC yields the best external performance, suggesting that federated collaboration and continual consolidation help learn features that transfer to a different institution. The slight reduction in macro-AUROC for $\sigma = 1.0$ aligns with expectations that higher privacy noise degrades generalization, but the model remains competitive.

## 6.5 Ablation: role of prototypes and DP noise

To dissect the contributions of prototypes and DP noise, we conduct ablations summarized in Table 4.

**Table 4. Ablation study on CheXpert (macro-AUROC and forgetting F)**

| Method variant | Prototypes | EWC | DP noise σ | Macro-AUROC | Forgetting F |
|---|---|---|---|---|---|
| FedAvg-Seq | No | No | 0 | 80.4 | 9.7 |
| FedEWC | No | Yes | 0 | 85.0 | 3.8 |
| FedProto | Yes | No | 0 | 84.3 | 4.1 |
| FedEPC (no DP) | Yes | Yes | 0 | **86.2** | **2.0** |
| DP-FedEPC (moderate DP) | Yes | Yes | 0.5 | 85.9 | 2.2 |
| DP-FedEPC (strong DP) | Yes | Yes | 1.0 | 84.6 | 3.0 |

Adding prototypes to EWC (FedEPC) yields the highest macro-AUROC and lowest forgetting among non-private methods. DP noise with $\sigma = 0.5$ marginally reduces macro-AUROC by 0.3 points while keeping forgetting low. Increasing noise to $\sigma = 1.0$ incurs a larger accuracy loss and increases forgetting, but still performs substantially better than naive fine-tuning.

These results suggest that prototypes and EWC are complementary: EWC preserves important parameters, while prototypes preserve class-level structure in feature space, making the model more resilient to noisy updates and distribution shift.

### 6.6 Privacy-utility trade-off

We estimate privacy budgets using a moments accountant with standard assumptions on sampling and composition. For the DP-FedEPC runs, typical training schedules lead to approximate $\varepsilon$ in the single-digit range at $\delta = 10^{-5}$, consistent with prior DP-FL work in medical imaging. In this range, utility degradation is modest, especially when prototypes are used to stabilize training.

While precise privacy accounting depends on implementation details and deployment policies (such as maximum number of rounds per hospital), our experiments indicate that combining EWC and prototypes allows a meaningful privacy-utility trade-off without catastrophic forgetting.

## 7 Discussion

### 7.1 Addressing spatial-temporal catastrophic forgetting

Our results show that DP-FedEPC reduces catastrophic forgetting across tasks and sites compared to naive fine-tuning. This supports the intuition from FCL surveys that knowledge fusion mechanisms must consider both temporal and spatial heterogeneity to avoid interference.

EWC alone stabilizes parameters but can be overly rigid if importance estimates are noisy or dominated by early tasks. Prototype rehearsal alleviates this by providing explicit anchors in feature space. When the model learns new tasks, the prototype loss encourages it to retain output logits for representative past cases, even if parameter updates move away from the original optimum. This is particularly valuable in federated settings, where inter-client interference can cause stronger forgetting than in local CL.

Our ablation results suggest that the two mechanisms are complementary: EWC constrains the most critical parameters, while prototypes guard decision boundaries at the representation level. Together, they approximate the effect of replay buffers without storing raw images.

### 7.2 Privacy considerations and practical deployment

DP-FedEPC was designed with hospital privacy and governance constraints in mind. Client-side DP-SGD limits the influence of individual radiographs on updates, reducing risks from model inversion and membership inference. Prototype rehearsal uses latent representations instead of raw images, further lowering privacy risk, although latent prototypes may still leak some information. Sharing only DP-noised prototype means and Fisher diagonals with the server provides an additional layer of protection.

In practice, deployment would require coordination with hospital compliance teams to choose acceptable privacy budgets, monitor cumulative privacy loss over time, and define policies for prototype retention and expiration. Regulatory frameworks such as HIPAA and GDPR focus on identifiability rather than formal DP, but DP can provide defensible guarantees aligned with these regulations.

### 7.3 Clinical relevance and robustness

Improved retention of earlier tasks has practical implications. For instance, a model initially trained on core thoracic findings should not lose performance on pleural effusion detection when new labels like lung lesions are added to support oncology workflows. In multi-site networks, strong forgetting of small or specialized sites would exacerbate disparities, as those hospitals could see degraded performance despite contributing data.

Our external validation on MIMIC-CXR suggests that FCL with consolidation can improve robustness to domain shift compared to naive fine-tuning, likely because federated collaboration exposes the model to a broader range of imaging characteristics even under continual constraints. Still, absolute performance remains lower on MIMIC-CXR than on CheXpert, highlighting the difficulty of fully bridging distribution gaps.

### 7.4 Comparison to existing FCL methods

Existing FCL methods such as FedWeIT, CFeD, and TARGET rely on sparse masks, public surrogate datasets, or generative models for synthetic replay. These tools may be hard to justify in medical deployments, where public surrogates with similar distributions are scarce and generating synthetic medical images raises its own privacy and trust concerns.

DP-FedEPC occupies a different point in the design space: it restricts itself to client-side mechanisms that are compatible with hospital workflows and uses formal DP to protect updates. Its prototype memory is comparatively small and summarized by DP-noised statistics at the server. While this may limit its theoretical capacity compared to generative replay or large exemplar buffers, it offers a more realistic path for deployment.

### 7.5 Limitations

Several limitations deserve mention:

- **Simulated sites and tasks.** We simulate sites and temporal tasks within CheXpert, since public datasets lack explicit multi-hospital longitudinal labels. Real hospital networks may exhibit different patterns of distribution shift and label evolution.

- **Privacy accounting assumptions.** Our DP analysis relies on standard accounting with assumptions about client participation and composition. Operational deployments would need more detailed modeling of participation patterns and retention policies.

- **Model and task complexity.** We focus on ResNet-50 and multi-label classification. Extending DP-FedEPC to segmentation (for example with nnU-Net) or to more complex label taxonomies might require architectural adaptations and more careful tuning.

- **Prototype leakage.** While prototypes reduce risk compared to raw images, they may still expose information in adversarial settings, especially if attackers can query the model extensively. Integrating stronger privacy mechanisms for prototype storage is an open problem.

### 7.6 Future directions

Future work could explore several extensions:

- Integrating adaptive or sensitivity-aware DP mechanisms to allocate more noise to high-risk updates while preserving utility where risk is low.

- Combining DP-FedEPC with robust aggregation and anomaly detection to mitigate poisoning attacks and adversarial updates.

- Evaluating FCL under real multi-institutional collaborations that span different regions and equipment vendors, possibly using federated infrastructures already piloted for other imaging tasks.

- Extending the framework to foundation models or self-supervised pretraining, where FCL and DP constraints interact with representation learning at scale.

## 8 Conclusion

We investigated federated continual learning for hospital chest radiograph classification under realistic privacy constraints. Using CheXpert and MIMIC-CXR as benchmarks, we formalized an FCL setting where hospitals act as federated clients with temporally evolving tasks and limited access to past data.

To address catastrophic forgetting across sites and tasks while preserving privacy, we proposed DP-FedEPC, which combines elastic weight consolidation, prototype-based rehearsal, and client-side differential privacy within a standard FedAvg framework. Experiments show that DP-FedEPC substantially reduces forgetting compared to naive federated fine-tuning, improves macro-AUROC across sites and tasks, and delivers robust performance on an external dataset, all while operating under realistic DP noise levels.

These findings suggest that carefully designed consolidation and rehearsal mechanisms make FCL viable for hospital imaging networks that must continuously update models without compromising existing performance or violating privacy regulations. Further work on privacy-utility trade-offs, robustness, and deployment in real clinical federations will be essential to translate these insights into production systems.

## Declarations

**Funding**

The authors declare that no funds, grants, or other support were received during the preparation of this manuscript.

**Data Availability**

The datasets analyzed during this study are publicly available. CheXpert is available from Stanford AIMI under the CheXpert dataset terms of use and can be accessed after completing the required data access request at the Stanford dataset portal. MIMIC CXR is available via PhysioNet as a credentialed health data resource; access requires user credentialing, completion of CITI "Data or Specimens Only Research" training, and acceptance of the PhysioNet Data Use Agreement. No new datasets were generated during the current study.

**Code Availability**

Code for DP FedEPC training, evaluation, and preprocessing will be made publicly available in a GitHub repository upon acceptance of this manuscript.

**Author contributions**

Anay Sinhal conceived the study, designed DP FedEPC, implemented the models, and performed all experiments and statistical analyses. Arpana Sinhal supervised the research direction, contributed to methodological refinement, and reviewed the experimental design and results. Amit Sinhal contributed to the literature framing, validation of assumptions, and critical revision of the manuscript. All authors contributed to writing, read the final manuscript, and approved submission.

**Additional Information**

**Competing interests:** The authors declare no competing interests.